\pdfoutput=1

\documentclass[11pt]{article}

\usepackage{acl}

\usepackage{times}
\usepackage{latexsym}

\usepackage[T1]{fontenc}

\usepackage[utf8]{inputenc}

\usepackage{microtype}
\usepackage{multirow}
\usepackage{graphicx}
\usepackage{enumerate} 
\usepackage{caption}
\usepackage{subcaption}
\usepackage{xurl}
\usepackage{xcolor}
\usepackage{amssymb}

\usepackage{booktabs}
\usepackage{adjustbox}
\usepackage{tabularx}

\definecolor{darkturquoise}{rgb}{0.0, 0.81, 0.82}
\definecolor{bondiblue}{rgb}{0.11, 0.67, 0.84}
\definecolor{dandelion}{rgb}{0.94, 0.88, 0.19}
\definecolor{darkmagenta}{rgb}{0.55, 0.0, 0.55}
\definecolor{darkraspberry}{rgb}{0.53, 0.15, 0.34}
\definecolor{darkpink}{rgb}{0.91, 0.33, 0.5}
\definecolor{darkred}{rgb}{0.55, 0.0, 0.0}

\title{Findings of the VarDial Evaluation Campaign 2023}

\author{No{\"e}mi Aepli\textsuperscript{1}, Çağrı Çöltekin\textsuperscript{2}, Rob van der Goot\textsuperscript{3},  Tommi Jauhiainen\textsuperscript{4} \\ [2pt]
\bf Mourhaf Kazzaz\textsuperscript{2}, Nikola Ljubešić\textsuperscript{5,6},
\bf  Kai North\textsuperscript{7}, Barbara Plank\textsuperscript{8} \\ [2pt] 
\bf Yves Scherrer\textsuperscript{4},  Marcos Zampieri\textsuperscript{7}\\[8pt]
  \textsuperscript{1}University of Zurich,
  \textsuperscript{2}University of Tübingen,
  \textsuperscript{3}IT University of Copenhagen,\\[2pt]
  \textsuperscript{4}University of Helsinki,
  \textsuperscript{5}Jožef Stefan Institute, 
  \textsuperscript{6}University of Zagreb,  \\[2pt]
  \textsuperscript{7}George Mason University,
  \textsuperscript{8}LMU Munich}

\begin{document}
\maketitle
\begin{abstract}

This report presents the results of the shared tasks organized as part of the VarDial Evaluation Campaign 2023. The campaign is part of the tenth workshop on Natural Language Processing (NLP) for Similar Languages, Varieties and Dialects (VarDial), co-located with EACL 2023. Three separate shared tasks were included this year: Slot and intent detection for low-resource language varieties (SID4LR), Discriminating Between Similar Languages -- True Labels (DSL-TL), and Discriminating Between Similar Languages -- Speech (DSL-S). All three tasks were organized for the first time this year. 
\end{abstract}

\section{Introduction}


The workshop series on \textit{NLP for Similar Languages, Varieties and Dialects} (VarDial), traditionally co-located with international conferences, has reached its tenth edition. Since the first edition, VarDial has hosted shared tasks on various topics such as language and dialect identification, morphosyntactic tagging, question answering, and cross-lingual dependency parsing. The shared tasks have featured many languages and dialects from different families and data from various sources, genres, and domains \cite{aepli-etal-2022-findings,chakravarthi-etal-2021-findings-vardial, gaman-etal-2020-report,zampieri-etal-2019-report, zampieri-etal-2018-language, zampieri-etal-2017-findings, malmasi-etal-2016-discriminating, zampieri-etal-2015-overview, zampieri-etal-2014-report}.

As part of the VarDial Evaluation Campaign 2023, we offered three shared tasks which we present in this paper:

\begin{itemize}
\item \textbf{SID4LR:} Slot and intent detection for low-resource language varieties\footnote{Task organizers: No{\"e}mi Aepli, Rob van der Goot, Barbara Plank, Yves Scherrer.}
\item \textbf{DSL-TL:} Discriminating Between Similar Languages -- True Labels\footnote{Task organizers: Marcos Zampieri, Kai North, Tommi Jauhiainen.}
\item \textbf{DSL-S:} Discriminating Between Similar Languages -- Speech\footnote{Task organizers: Çağrı Çöltekin, Mourhaf Kazzaz, Tommi Jauhiainen, Nikola Ljubešić.}
\end{itemize}

DSL-TL and DSL-S continue the long line of language and dialect identification \cite{jauhiainen2019survey} shared tasks at VarDial, whereas the SID4LR features a task novel to the evaluation campaigns.

This overview paper is structured as follows: in Section~\ref{sec:tasks}, we briefly introduce the three shared tasks. Section~\ref{sec:participants} presents the teams that submitted systems to the shared tasks. Each task is then discussed in detail, focusing on the data, the participants' approaches, and the obtained results. Section~\ref{sec:SID4LR} is dedicated to SID4LR, Section~\ref{sec:dsltl} to DSL-TL, and Section~\ref{sec:dsls} to DSL-S.



\section{Shared Tasks at VarDial 2023} \label{sec:tasks}

The evaluation campaign took place in January -- February 2023. Due to the ACL placing the workshop at the EACL conference in early May, the schedule from the shared tasks' first announcement to completion was relatively tight. The call for participation in the shared tasks was first published in early January, the training data sets for the shared tasks were released on January 23\textsuperscript{rd}, and the results were due to be submitted on February 27\textsuperscript{th}.\footnote{\url{https://sites.google.com/view/vardial-2023/shared-tasks}}

\subsection{SID for Low-resource Language Varieties (SID4LR)}\label{sec:SID4LR-intro}

\begin{figure*}
\centering
\begin{tabular}{ll}
English (EN)  & Remind me to \textcolor{darkturquoise}{go to the dentist} \textcolor{darkpink}{next Monday} \\
Italian (IT) & Ricordami di \textcolor{darkturquoise}{andare dal dentista} \textcolor{darkpink}{lunedì prossimo} \\
\textbf{Neapolitan (NAP)} & \textbf{Ricuordam' `e \textcolor{darkturquoise}{`i addo dentista} \textcolor{darkpink}{lunnerì prossimo}}      \\
German (DE) & Erinnere mich am \textcolor{darkpink}{nächsten Montag} zum \textcolor{darkturquoise}{Zahnarzt zu gehen} \\
\textbf{Swiss German (GSW)} & \textbf{Du mi dra erinnere \textcolor{darkpink}{nöchscht Mänti} zum \textcolor{darkturquoise}{Proffumech zga}} \\
\textbf{South Tyrolean (DE-ST)}  & \textbf{Erinner mi in \textcolor{darkpink}{negschtn Muntig} \textcolor{darkturquoise}{zin Zohnorzt zu gian}}   \\
         
\end{tabular}
\caption{Example of the SID tasks. The \textbf{three target languages (NAP, GSW, DE-ST)} are in bold, the corresponding high-resource languages (DE and IT) and the translation (EN) are included for comparison. The \textit{slot} annotations are coloured: \textcolor{darkpink}{datetime} and \textcolor{darkturquoise}{reminder/todo}. The \textit{intent} for this sentence is \texttt{reminder/set\_reminder}.}
\label{tab:sid4lrexample}
\end{figure*}


The SID4LR shared task focused on Slot and Intent Detection (SID) for digital assistant data in three low-resource language varieties: Swiss German (GSW) from the city of Bern, South Tyrolean (DE-ST), and Neapolitan (NAP). Intent detection is the task of automatically classifying the intent of an utterance and slot detection aims at finding the relevant (labeled) span. Figure \ref{tab:sid4lrexample} illustrates these two tasks with an example. The objective of this shared task is to address the following question: \textit{How can we best do zero-shot transfer to low-resource language varieties without standard orthography?}

The xSID-0.4 corpus\footnote{\url{https://bitbucket.org/robvanderg/sid4lr}}, which includes data from both Snips \citep{coucke_snips_2018} and Facebook \citep{schuster-etal-2019-cross-lingual}, constitutes the training data, providing labeled information for slot and intent detection in 13 different languages. The original training data is in English, but we also provided automatic translations of the training data into German, Italian, and other languages. These translations are obtained with the Fairseq library~\cite{ott-etal-2019-fairseq}, using spoken data for training (more details in~\newcite{van-der-goot-etal-2021-masked}). Bleu scores~\cite{papineni-etal-2002-bleu} were 25.93 and 44.73 for respectively German and Italian. Slot label annotations were transferred using the attention weights. Participants were allowed to use other data to train on as long as it was not annotated for SID in the target languages. Specifically, the following resources were allowed:

\begin{enumerate}
\item annotated data from other (related and unrelated) languages in the xSID-0.4 corpus;
\item raw text data from the target languages, if available (e.g., Wikipedia, web crawls);
\item pre-trained language models containing data from the target languages.
\end{enumerate}

It was not mandatory for the participants to provide systems for all tasks and languages; they had the option to only take part in a specific subset.
We used the standard evaluation metrics for these tasks, namely the span F1 score for slots and accuracy for intents.

\subsection{Discriminating Between Similar Languages -- True Labels (DSL-TL)}
\label{sec:dsltl-intro}

Discriminating between similar languages (e.g., Croatian and Serbian) and national language varieties (e.g., Brazilian and European Portuguese) has been a popular topic at VarDial since its first edition. The DSL shared tasks organized from 2014 to 2017 \cite{zampieri-etal-2017-findings, malmasi-etal-2016-discriminating, zampieri-etal-2015-overview, zampieri-etal-2014-report} have addressed this issue by providing participants with the DSL Corpus Collection (DSLCC) \cite{Tan2014}, a collection of journalistic texts containing texts written in groups of similar languages (e.g., Indonesian and Malay) and language varieties (e.g., Brazilian and European Portuguese).\footnote{\url{http://ttg.uni-saarland.de/resources/DSLCC/}} The DSLCC was compiled assuming each instance's gold label is determined by where the text is retrieved from. While this is a straightforward and primarily accurate practical assumption, previous research \cite{goutte-etal-2016-discriminating} has shown the limitations of this problem formulation as some texts may present no linguistic marker that allows systems or native speakers to discriminate between two very similar languages or language varieties.

At VarDial 2023, we tackle this important limitation by introducing the DSL True Labels (DSL-TL) shared task. DSL-TL provided participants with the DSL-TL dataset \cite{zampieri2023language}, the first human-annotated language variety identification dataset where the sentences can belong to several varieties simultaneously. The DSL-TL dataset contains newspaper texts annotated by multiple native speakers of the included language and language varieties, namely English (American and British varieties), Portuguese (Brazilian and European varieties), and Spanish (Argentinian and Peninsular varieties). More details on the DSL-TL shared task and dataset are presented in Section \ref{sec:dsltl}.

\begin{table*}
\centering
\begin{tabular}{l|cccl}
\textbf{Team} & \textbf{SID4LR} &  \textbf{DSL-TL} &  \textbf{DSL-S} &  \textbf{System Description Paper}\\
\midrule
UBC &  \checkmark &  & & \citet{2023-ubc} \\
Notre Dame & \checkmark & & &  \citet{2023-notredame} \\
VaidyaKane & & \checkmark & &  \citet{2023-vaidyakane} \\
ssl & & \checkmark & &  \citet{2023-ssl} \\
UnibucNLP & & \checkmark & &  \citet{2023-UnibucNLP} \\
SATLab & & \checkmark & & \\
\end{tabular}
\caption{\label{tab:overview}
The teams that participated in the VarDial Evaluation Campaign 2023.
}
\end{table*}

\subsection{Discriminating Between Similar Languages -- Speech (DSL-S)}\label{sec:dsls-intro}

In the DSL-S 2023 shared task, participants were using the training, and the development sets from the Mozilla Common Voice \cite[CV,][]{ardila-etal-2020-common} to develop a language identifier for speech.\footnote{Further information available at: \url{https://dsl-s.github.io}.} The nine languages selected for the task come from four different subgroups of Indo-European or Uralic language families (Swedish, Norwegian Nynorsk, Danish, Finnish, Estonian, Moksha, Erzya, Russian, and Ukrainian).

The 9-way classification task was divided into two separate tracks. Only the training and development data from the CV dataset were allowed in the closed track, and no other data were to be used. This prohibition included systems and models trained (unsupervised or supervised) on any other data. On the open track, the use of any openly available (available to any possible shared task participant) datasets and models not including or trained on the Mozilla Common Voice test set was allowed. The evaluation measure used was the Macro F1 score over the nine languages.

\section{Participating Teams} \label{sec:participants}

A total of six teams submitted runs to the SID4LR and DSL-TL tasks. Two teams registered for the DSL-S shared task, but neither provided any submissions. In Table~\ref{tab:overview}, we list the teams that participated in the shared tasks, including references to the system description papers, which are published as parts of the VarDial workshop proceedings. Detailed information about the submissions is included in the task-specific sections below.

\section{SID for Low-resource Language Varieties} \label{sec:SID4LR}

\subsection{Dataset}

The xSID-0.4 corpus\footnote{\url{https://bitbucket.org/robvanderg/sid4lr}} makes up the training data and provides labeled information for slot and intent detection in 13 different languages. The xSID dataset consists of sentences from the English Snips \citep{coucke_snips_2018} and cross-lingual Facebook \citep{schuster-etal-2019-cross-lingual} datasets, which were manually translated into 12 other languages \cite{van-der-goot-etal-2021-masked}.
There are 43,605 sentences in the English training data. The evaluation data contains 500 test sentences and 300 validation sentences per language.
For the test data, we took the existing South Tyrolean (DE-ST) part of xSID \citep{van-der-goot-etal-2021-masked} and two novel translations created for this shared task: Bernese Swiss German (GSW) and Neapolitan (NAP). 
The new translations were done by native speakers of the two language varieties. They translated directly from English without seeing the Italian or German source sentences.
The translations were then processed and annotated by the shared task organizers (who have passive knowledge of the two language varieties). The two steps were done according to the guidelines from the original paper by \citet{van-der-goot-etal-2021-masked}.

\subsection{Participants and Approaches}\label{sec:SID4LR_approaches}

\paragraph{UBC:} Team UBC \citep{2023-ubc} participated in both subtasks: slot and intent detection. They used several multilingual Transformer-based language models, including mBERT, XLM-R, SBERT, LaBSE, LASER, and mT0. Furthermore, they experimented with a variety of settings to improve performance: varying the source languages, combining different language models, data augmentation via paraphrasing and machine translation, and pre-training on the target languages. For the latter, they made use of additional external data from various sources for all three target languages for the training.

\paragraph{Notre Dame:} Team Notre Dame \citep{2023-notredame} submitted a research paper to the VarDial workshop, within which they also described their participation in the intent detection subtask. The team applied zero-shot methods, i.e., they did not use any data from the target language in the training process.
They fine-tuned monolingual language models\footnote{German BERT: \url{https://huggingface.co/dbmdz/bert-base-german-uncased} and Italian BERT: \url{https://huggingface.co/dbmdz/bert-base-italian-uncased}} with noise-induced data. The noising technique they applied is similar to that of \citet{aepli-sennrich-2022-improving} with three main differences: they 1) add an additional noise type: \textit{swapping} between adjacent letters; 2) they employ higher levels of noise and include multiple copies of the fine-tuning data; and 3) remove the step of continued pre-training to avoid using any target language data.

\paragraph{Baseline:} The baseline we provided is the same as in the original xSID paper, trained on the English data, with an updated version of MaChAmp \citep{van-der-goot-etal-2021-massive}. The model uses an mBERT encoder and a separate decoder head for each task, one for slot detection (with a CRF layer) and one for intent classification.

\subsection{Results}

We evaluated the submitted systems according to accuracy for intents and according to the span F1 score for slots (where both span and label must match exactly). Table \ref{tab:sid4lr_results} contains the scores.

For intent classification, the winner for all three languages is the team Notre Dame. Both teams beat the baseline by a large margin. All systems reached the highest scores on DE-ST and the lowest scores on GSW, but both participating teams managed to significantly close the gaps between the languages compared to the baseline. 

For slot detection, the UBC team outperformed the baseline for DE-ST and GSW but not for NAP. Again, GSW turned out to be the most difficult language variety of the three. We must note, however, that the UBC submission contained a large amount of ill-formed slots. Between 13\% (DE-ST, NAP) and 28\% (GSW) of predicted slots start with an \texttt{I-} label instead of \texttt{B-}; the evaluation script simply ignores such slots. Furthermore, a small number of predicted spans have inconsistent labels (e.g., \texttt{I-datetime} immediately followed by \texttt{I-location}). This suggests that the model architecture chosen by the UBC team was not appropriate for span labeling tasks and that a different architecture could have led to further improvements compared to the baseline. The baseline system, which uses a CRF prediction layer, did not produce any such inconsistencies.

\begin{table}[ht]
\centering
\resizebox{\columnwidth}{!}{%
\begin{tabular}{ll|ccc}
\textbf{}                                                                            & \textbf{}   & \textbf{Baseline} & \textbf{UBC}    & \textbf{Notre Dame} \\ \midrule
\multirow{3}{*}{\textbf{\begin{tabular}[c]{@{}l@{}}Intent\\ detection\end{tabular}}} & \textbf{DE-ST} & 0.6160            & 0.8940          & \textbf{0.9420}     \\
 & \textbf{GSW} & 0.4720 & 0.8160          & \textbf{0.8860} \\
 & \textbf{NAP} & 0.5900 & 0.8540          & \textbf{0.8900} \\
 \midrule
\multirow{3}{*}{\textbf{\begin{tabular}[c]{@{}l@{}}Slot\\ detection\end{tabular}}}   & \textbf{DE-ST} & 0.4288            & \textbf{0.4692} &  --                   \\
 & \textbf{GSW} & 0.2530 & \textbf{0.2899} &      --           \\
 & \textbf{NAP} & \textbf{0.4457} & 0.4215 &      --          \\ 

\end{tabular}%
}
\caption{Results for intent classification (accuracy) and slot detection (Span-F1 score). UBC submitted several models for intent detection, and here we report their best-performing system for each language.}
\label{tab:sid4lr_results}
\end{table}

\subsection{Summary}

The UBC submissions are based on a pre-trained multilingual language model (mT0), which was fine-tuned on the 12 languages of the xSID dataset. Among these languages are Italian and German, but all training sets except the English one have been produced by machine translation. This setup worked better than using only the related languages of xSID (IT and DE) or only English. Also, further data augmentation with paraphrasing and machine translation did not have any positive effect. These findings suggest that task-specific knowledge is more important than having access to linguistic material in the target languages (or even in related high-resource languages).

The Notre Dame participation provides a somewhat contrasting result. They start with a monolingual BERT model of the related high-resource language (IT or DE) and use fine-tuning to make the model more robust to character-level noise. The possibility of including unrelated languages was not explored here.

The contributions proposed by the participants are thus largely complementary, and it would be interesting to see if their combination leads to further improvements on the task. For instance, task-specific fine-tuning (using all of the xSID data) could be combined with language-specific fine-tuning (based on the noise induction task) and complemented with the baseline's CRF architecture to provide consistent slot labels. 


A striking finding of this shared task are the poor results on Swiss German compared to the other two low-resource varieties, Neapolitan and South-Tyrolean German. This may be due to the particular Swiss German dialect used in this dataset and/or to some translator-specific preferences or biases. Further analysis will be required to fully explain these differences.

 \begin{table*}
\centering
\scalebox{0.90}{\begin{tabular}{l|ccc|c}

     \textbf{Language} & \textbf{Variety A} &  \textbf{Variety B} &  \textbf{Both/Neither} & \bf Total \\ 
     \hline
     Portuguese & 1,317 (pt-PT) & 3,023 (pt-BR) & 613 (pt) & 4,953\\
     Spanish & 2,131 (es-ES) & 1,211 (es-AR) & 1,605 (es) & 4,947\\
     English & 1,081 (en-GB) & 1,540 (en-US) & 379 (en) & 3,000\\
 \hline
\bf Total & & & & \bf 12,900 \\
\end{tabular}}
 \caption{\label{dataset_table1} DSL-TL's class splits and the total number of instances.}
\end{table*}

\section{Discriminating Between Similar Languages -- True Labels} \label{sec:dsltl}

The DSL-TL shared task contained two tracks:

\begin{itemize}
    \item {\bf Track 1 -- Three-way Classification:} In this track, systems were evaluated with respect to the prediction of all three labels for each language, namely the variety-specific labels (e.g., PT-PT or PT-BR) and the common label (e.g., PT).
    \item {\bf Track 2 -- Binary Classification:} In this track, systems were scored only on the variety-specific labels (e.g., EN-GB, EN-US).
\end{itemize}

\noindent In addition to the two tracks mentioned above, we provided participants with the option of using external data sources (open submission) or only the DSL-TL dataset (closed submission).

\subsection{Dataset}

\paragraph{Data} DSL-TL contains 12,900 instances split between three languages and six national language varieties, as shown in Table \ref{dataset_table1}. Instances in the DSL-TL are short extracts (1 to 3 sentences long) from newspaper articles randomly sampled from two sources \cite{zellers2019defending,Tan2014}. Considering the source's ground truth label, the DSL-TL creators randomly selected 2,500 instances for each Portuguese and Spanish variety and 1,500 instances for each English variety.

\paragraph{Annotation} DSL-TL was annotated using crowdsourcing through Amazon Mechanical Turk (AMT).\footnote{\url{https://www.mturk.com/}} The annotation task was restricted to annotators based on the six national language variety countries, namely Argentina, Brazil, Portugal, Spain, United Kingdom, and the United States. The annotators were asked to label each instance with what they believed to be the most representative variety label, namely European (pt-PT) or Brazilian Portuguese (pt-BR), Castilian (es-ES) or Argentine Spanish (es-AR), and British (en-GB) or American English (en-US). The label distributions are shown in Table \ref{dataset_table1}. The annotators were presented with three choices: (1) language variety A, (2) language variety B, or (3) both or neither for cases in which no clear language variety marker (either linguistic or named entity) was present in the text. The annotator agreement calculations and filtering carried out after the annotation stage are described in detail in the dataset description paper \cite{zampieri2023language}. Finally, the instances in DSL-TL have been split into training, development, and testing partitions, as shown in Table \ref{train_test_splits}. 

 \begin{table}[t]
\centering
\scalebox{0.92}{\begin{tabular}{l|ccc|c}
     \textbf{Variety} & \textbf{Train} &  \textbf{Dev} &  \textbf{Test} & \bf Total \\ 
     \hline
     Portuguese & 3,467 & 991 & 495 & 4,953\\
     Spanish & 3,467 & 985 & 495 & 4,947\\  
     English & 2,097 & 603 & 300 & 3,000\\ 
 \hline
 Total & & & & 12,900 \\
\end{tabular}}
 \caption{\label{train_test_splits} DSL-TL's train, dev, and test splits are 70/20/10\% of the total number of instances, respectively.}
\end{table}

\subsection{Participants and Approaches}

Four teams provided submissions to the shared task.

\paragraph{VaidyaKane:} All submissions from the team VaidyaKane used a pre-trained multilingual XLM-RoBERTa fine-tuned to language identification\footnote{\url{https://huggingface.co/papluca/xlm-roberta-base-language-detection}} to classify the language of the sentence \cite{conneau-etal-2020-unsupervised}. After the initial language identification, they experimented with several language-specific BERT models to identify the exact variety. Their best submission on track one used ``bert-base-uncased''\footnote{\url{https://huggingface.co/bert-base-uncased}} for English \cite{devlin-etal-2019-bert}, ``bertin-project/bertin-roberta-base-spanish''\footnote{\url{https://huggingface.co/bertin-project/bertin-roberta-base-spanish}} for Spanish \cite{BERTIN}, and ``neuralmind/bert-base-portuguese-cased''\footnote{\url{https://huggingface.co/neuralmind/bert-base-portuguese-cased}} for Portuguese \cite{souza2020bertimbau}. On track two, the models for Spanish and Portuguese were the same, but ``roberta-base''\footnote{\url{https://huggingface.co/roberta-base}} was used for English \cite{DBLP:journals/corr/abs-1907-11692}.

\paragraph{ssl:} Team ssl submitted one submission to each of the four track combinations. For the closed tracks, they trained an SVM classifier using TF-IDF weighted character n-grams from one to four and word n-grams from one to two. On the open tracks, they also used names of people obtained from Wikidata \cite{10.1145/2629489}.

\paragraph{UnibucNLP:} On track one, the UnibucNLP team submitted a run using an XGBoost stacking ensemble \cite{10.1145/2939672.2939785}. The classifier stack for the ensemble consisted of one SVM and one KRR classifier. For track two, the stack classifiers were the same, but Logistic Regression was used for the stacking ensemble.

\paragraph{SATLab:} On both tracks, the SATLab team used a Logistic Regression classifier from the LIBLinear package with character n-grams from one to five weighted by BM25 and L2 normalization. The n-grams had to appear in at least two different sentences in the training data. The system was very similar to the one used by \citet{bestgen-2021-optimizing} in the Dravidian Language Identification (DLI) shared task in 2021 \cite{chakravarthi-etal-2021-findings-vardial}.

\subsection{Results}

Tables \ref{tab:testing1} to \ref{tab:testing4} show the recall, precision, and F1 scores for the baselines and best submissions for all track combinations.

\begin{table}[!ht]\centering
\scalebox{.80}{
\begin{tabular}{c|lccc}
\bf Rank & \bf Model & \bf R & \bf P & \bf F1 \\ 
\hline 
 & baseline-mBERT & 0.5490 & 0.5450 & 0.5400 \\ 
 & baseline-XLM-R & 0.5280 & 0.5490 & 0.5360 \\ 
1 & run-3-UnibucNLP & 0.5291 & 0.5542 & 0.5318 \\ 
 & baseline-NB & 0.5090 & 0.5090 & 0.5030 \\ 
2 & run-1-SATLab & 0.4987 & 0.4896 & 0.4905 \\ 
3 & run-1-ssl & 0.4978 & 0.4734 & 0.4817 \\  
\end{tabular}
}
\caption{The macro average scores of the best run for each team on \textbf{closed track 1}.}
\label{tab:testing1}
\end{table}

\begin{table}[!ht]\centering
\scalebox{.80}{
\begin{tabular}{c|lccc}
\bf Rank & \bf Model & \bf R & \bf P & \bf F1 \\ 
\hline 
 & baseline-ANB & 0.8200 & 0.7990 & 0.7990 \\ 
 & baseline-NB & 0.8110 & 0.7920 & 0.7940 \\ 
 & baseline-XLM-R & 0.7830 & 0.7820 & 0.7800 \\ 
1 & run-1-ssl & 0.7521 & 0.7885 & 0.7604 \\ 
 & baseline-mBERT & 0.7600 & 0.7530 & 0.7550 \\ 
2 & run-2-SATLab & 0.7520 & 0.7430 & 0.7452 \\ 
3 & run-1-UnibucNLP & 0.6502 & 0.7756 & 0.6935 \\  
\end{tabular}
}
\caption{The macro average scores of the best run for each team on \textbf{closed track 2}.}
\label{tab:testing2}
\end{table}

\begin{table}[!ht]\centering
\scalebox{.82}{
\begin{tabular}{c|lccc}
\bf Rank & \bf Model & \bf R & \bf P & \bf F1 \\ \hline 
1 & run-3-VaidyaKa & 0.5962 & 0.5866 & 0.5854 \\ 
2 & run-1-ssl & 0.4937 & 0.5068 & 0.4889 \\  
\end{tabular}
}
\caption{The macro average scores of the best run for \textbf{open track 1}.}
\label{tab:testing3}
\end{table}

\begin{table}[!ht]\centering
\scalebox{.82}{
\begin{tabular}{c|lccc}
\bf Rank & \bf Model & \bf R & \bf P & \bf F1 \\ 
\hline 
1 & run-1-VaidyaKa & 0.8705 & 0.8523 & 0.8561 \\ 
 & baseline-NB & 0.8200 & 0.8030 & 0.8030 \\ 
2 & run-1-ssl & 0.7647 & 0.7951 & 0.7729 \\  
\end{tabular}
}
\caption{The macro average scores of the best run for each team on \textbf{open track 2}.}
\label{tab:testing4}
\end{table}

Team UnibucNLP \cite{2023-UnibucNLP} achieved the first place out of nine submissions on the closed version of track one. Their XGBoost stacking ensemble attained 
an F1 score of 0.5318. The results were still slightly worse than the multilingual BERT\footnote{mBERT: \url{https://huggingface.co/bert-base-multilingual-cased}} (mBERT) \cite{devlin-etal-2019-bert} and the XLM-RoBERTa\footnote{XLM-R: \url{https://huggingface.co/xlm-roberta-base}}  (XLM-R) \cite{DBLP:journals/corr/abs-1907-11692} baselines. All other submissions achieved slightly worse F1 scores. In the second place, team SATLab's logistic regressor obtained an F1 score of 0.4905. In third place, team ssl's SVM produced an F1 score of 0.4817. The similarity between the top three F1 scores shows that automatically differentiating between similar language varieties is a challenging task, especially when taking into consideration neutral labels (EN, ES, or PT), as well as only using the provided data. 

Team ssl \cite{2023-ssl} achieved the best performance out of ten submissions on the closed version of track two. Their SVM was able to more effectively differentiate between six labels that did not include the aforementioned neutral labels (en-GB, en-US, es-AR, es-ES, pt-PT, or pt-BR). They achieved 
an F1 score of 0.7604. Their results were closely followed by the performance of SATLab's logistic regressor, having attained an F1 score of 0.7452, and UnibucNLP's XGBoost stacking ensemble with an F1 score of 0.6935. All submissions were clearly behind the adaptive and traditional Naive Bayes baselines, which were identical to the systems winning the Identification of Languages and Dialects of Italy (ITDI) shared task in 2022 \cite{jauhiainen-etal-2022-italian,aepli-etal-2022-findings}. SVMs are well-known to perform well when there is a clear distinction between class boundaries. This likely explains why team ssl's SVM has outperformed UnibucNLP's ensemble since neutral labels that contained features of both classes were no longer considered. 

Team VaidyaKane's \cite{2023-vaidyakane} submission to the open version of track 1 outperformed all other open and closed submissions for this track. Their two-stage transformer-based model achieved 
an F1 score of 0.5854. Team ssl was the only other team to submit predictions for open tracks 1 and 2. Their open submission for track 1 achieved an F1 score of 0.4889 which surpassed that of their closed submission for this track. The use of additional data was, therefore, found to improve overall performances.

Team VaidyaKane produced the highest F1 score on the open version of track 2. They achieved 
an F1 score of 0.8561, which was greater than all other open and closed submissions for either track. Team ssl also saw a further improvement in their SVM's model performance when using additional data for track 2. Their SVM model produced an F1 score of 0.7729, which was superior to their closed-track submission. These performances show that the use of additional data is beneficial and further proves that the classification of language varieties is an easier task than the classification of language varieties with neutral labels.

\subsection{Summary}

The DSL-TL shared task introduced a novel problem formulation in language variety identification. The new human-annotated dataset with the presence of the `both or neither' class represent a new way of looking at the problem. Given the similarity between language varieties, we believe this new problem formulation constitutes a fairer way of evaluating language identification systems, albeit rather challenging in terms of performance as demonstrated in this shared task. 

\section{Discriminating Between Similar Languages -- Speech} \label{sec:dsls}

\subsection{Dataset}

The DSL-S shared task uses Mozilla Common Voice data (version 12 released in Dec 2022)
in 9 languages from two language families.
The data comes from volunteers reading a pre-selected set of sentences
in each language.
The audio is recorded through a web-based interface.
For training and development sets, we follow the training and development set of the source data. Even though the test data used in this task comes from the Common Voice test data for the nine languages, we do not use the entire test set of the CV release
but sample 100 audio files for each language.
There is no overlap of sentences and speakers between the data sets.
Table~\ref{tbl:dsl-s-data} presents the test set's statistics.
The total amount of unpacked speech data is around 15 gigabytes. The data includes severe class imbalance,
as well as substantial differences in the number of speakers.
Generalization from a small number of speakers is a known challenge
in similar speech data sets,
including earlier VarDial evaluation campaigns.%
\footnote{See \citet{jauhiainen-etal-2018-heli-based} and \citet{wu-etal-2019-language}
for earlier approaches to this problem.}
The CV data set makes this task further challenging
since the variety of speakers in the test set is much larger
than the training and the development sets.

Similar to the earlier VarDial shared tasks
with audio data \cite{zampieri-etal-2017-findings,zampieri-etal-2018-language,zampieri-etal-2019-report},
we provided 400-dimensional i-vector and 512-dimensional x-vector features,
both extracted using Kaldi \cite{povey2011}.
Unlike earlier tasks, however, the raw audio data was also available to the potential participants.
\begin{table*}
  \centering
  \begin{tabular}{l|*{9}{r}}
    & \multicolumn{3}{c}{\textbf{Train}}&
      \multicolumn{3}{c}{\textbf{Dev}}&
      \multicolumn{3}{c}{\textbf{Test}}\\
    \cmidrule(lr){2-4} \cmidrule(lr){5-7} \cmidrule(lr){8-10}
    & n & spk & duration 
    & n & spk & duration 
    & n & spk & duration  \\
    \midrule
    \textbf{DA}  &2734 &3   &3:17:38   &2105   &10   &2:50:46  & 100 &48 &0:07:50\\
    \textbf{ET}  &3137 &221 &5:49:04   &2638   &167  &4:57:54  & 100 &88 &0:11:12\\
    \textbf{FI}  &2121 &3   &2:43:47   &1651   &13   &1:59:23  & 100 &63 &0:07:46\\
    \textbf{MDF} &173  &2   &0:15:39   &54     &1    &0:04:39  & 100 &7  &0:08:40\\
    \textbf{MYV} &1241 &2   &1:58:26   &239    &1    &0:22:55  & 100 &9  &0:09:07\\
    \textbf{NO}  &314  &3   &0:22:43   &168    &4    &0:13:28  & 100 &18 &0:07:35\\
    \textbf{RU}  &26043&252 &37:16:50  &10153  &394  &15:23:17 & 100 &98 &0:09:15\\
    \textbf{SV}  &7421 &22  &8:11:54   &5012   &73   &5:32:33  & 100 &89 &0:07:24\\
    \textbf{UK}  &15749&28  &18:38:31  &8085   &103  &10:58:25 & 100 &28 &0:08:22\\
  \end{tabular}
  \caption{Number of instances (n),
    number of speakers (spk) and total duration (hour:minute:seconds)
    for each split of the DSL-S shared task.
    The speaker numbers are approximated based on client id detection by CV.}%
    \label{tbl:dsl-s-data}
\end{table*}

\subsection{Participants and Approaches}

Two teams registered for the shared task, but neither provided any submissions.
In this section, we briefly introduce the baselines we provided.
For the closed track,
we provided a linear SVM baseline with x-vectors features \cite{snyder2018}.
The SVM baseline was implemented using scikit-learn \cite{sklearn},
and tuned only for the SVM margin parameter `C'.
The open track baseline uses two baselines - the XLS-R multilingual pre-trained transformer speech model~\cite{conneau2020unsupervised}\footnote{\url{https://huggingface.co/facebook/wav2vec2-large-xlsr-53}} with a classification head for direct speech classification, and  a multilingual speech recognition system
\footnote{\url{https://huggingface.co/voidful/wav2vec2-xlsr-multilingual-56}}
based on XLS-R~\cite{babu2021} to transcribe the speech,
and uses Naive Bayes \cite{jauhiainen-etal-2022-italian,jauhiainen-etal-2022-optimizing} to identify the language.\footnote{\url{https://github.com/tosaja/TunPRF-NADI}}

\subsection{Results}

\begin{table}
  \centering
  \begin{tabular}{l|rrr}
    \textbf{System} & \textbf{P} & \textbf{R} & \textbf{F1} \\
    \midrule
    SVM + x-vectors & 0.0914 & 0.1189 &  0.0876\\
    XLS-R & 0.6736 & 0.5953 & 0.5856 \\
    XLS-R + NB    & 0.7331 & 0.7167 & 0.7031\\
  \end{tabular}
  \caption{Baseline scores of the DSL-S shared task.}%
  \label{tbl:dsl-s-scores}
\end{table}

The scores for the baselines are presented in Table~\ref{tbl:dsl-s-scores}.
The SVM baseline performs particularly badly on the test set
(the development precision, recall, and F1 scores are
0.4088, 0.4011, 0.3777, respectively).
The reason behind this is likely due to the fact that,
although they were used for language identification in earlier research,
the x-vectors are designed for speaker identification.
Given the variability of speaker features in the test set,
any classifier relying on speaker features are likely to fail.
The baselines relying on pre-trained transformer models perform substantially better, with the direct speech classifier being more than 10 points behind the transcription and text classification approach. While the direct speech classification approach could be further improved through hyperparameter optimisation (currently we fine-tune for 3 epochs with a batch size of 24 and a learning rate of 1e-04) and a selection of the layer from which the features are extracted (related work suggests that lower transformer layers are more informative for discriminating between languages~\cite{bartley2023accidental}), these baseline results show that transcription and text classification might still be a shorter path to a reasonably performing system for discriminating between similar languages than direct speech classification.

\subsection{Summary}
Although we did not have any submissions for this shared task,
we believe that the task includes many interesting challenges.
Based only on our baseline results,
identifying languages from a limited amount of data
(without pre-trained speech models) seems challenging,
yet this is particularly interesting for low-resource settings
and for investigating differences and similarities for closely related language varieties.
We hope to see more interest in the community for language/dialect identification from speech.

\section{Conclusion}

This paper presented an overview of the three shared tasks organized as part of the VarDial Evaluation Campaign 2023: Slot and intent detection for low-resource language varieties (SID4LR), Discriminating Between Similar Languages -- True Labels (DSL-TL), and Discriminating Between Similar Languages -- Speech (DSL-S).

\section*{Acknowledgements}
We thank all the participants for their interest in the shared tasks. 

The work related to the SID4LR shared task has received funding from the Swiss National Science Foundation (project nos.~191934 and 176727) and ERC Grant 101043235. The work related to the DSL-TL and DSL-S shared tasks has received partial funding from the Academy of Finland (funding decision no.~341798). The work related to the DSL-S shared task has received funding from the Slovenian  Research Agency within the research project J7-4642 and the research programme P6-0411.

\bibliography{vardial2023,anthology,custom}
\bibliographystyle{acl_natbib}

\clearpage


\end{document}